# Automatic verbal aggression detection for Russian and American imageboards

Denis Gordeev[ab1]

[a]*National Research Nuclear University MEPhI (Moscow Engineering Physics Institute), Kashirskoe highway, 31 , Moscow, 115409, Russia*
[b]*Moscow State Linguistic University, Ostozhenka, 38, Moscow, 119034, Russia*

**Abstract**

The problem of aggression for Internet communities is rampant. Anonymous forums usually called imageboards are notorious for their aggressive and deviant behaviour even in comparison with other Internet communities. This study is aimed at studying ways of automatic detection of verbal expression of aggression for the most popular American (4chan.org) and Russian (2ch.hk) imageboards. A set of 1,802,789 messages was used for this study. The machine learning algorithm word2vec was applied to detect the state of aggression. A decent result is obtained for English (88%), the results for Russian are yet to be improved.

## 1. Introduction

The Internet is sometimes considered a quite violent and rude place. Many people, especially active users, face with cyberbullying and other expressions of aggression on a daily basis. For example, the U.S. Department of Health & Human Services has launched an initiative to stop bullying, including Internet bullying [1]. Imageboards that have been a buzzword for a while are considered a truly epicentre of all kind of unruly behaviours that we can find on the Net. No wonder they are called 'the Internet hate machine' [2].

Imageboards are usual Internet forums with no registration. Messages contain no personal details, only the message itself, date and email. However, registration mechanisms are not implemented and there is nothing that may prevent another person using the same information that you have provided with your message. Personal tripcodes are the only means to state your identity but they are used only in about 4% of cases [2]. It seems quite natural that aggression will flourish in such an environment where nobody can track you and where there are no social limits. Nevertheless, Potapova and Gordeev [3] have shown that it may be not true for Russian Internet communities, although the results are still disputed.

In this research, we study aggression in the environment where it is vividly presented and is not constrained by social boundaries. This research is important because it is one of the first works on automatic detection of verbal aggression. We also release our trained neural model that can be used for other researches and our methods may be used with some tweaking for other languages.

## *2*. Related works

Many researchers deal with aggression and its representation on the Internet. Potapova has been investigating aggression [4] and compiled a Russian dictionary containing words describing this emotional state [5]. Bernstein has conducted a research on 4chan and imageboard culture [2]. The task of sentiment analysis is rather close to aggression analysis because both tasks deal with detection of different human emotions. Twitter and social networks sentiment analysis are especially close to our research field, because the majority of anonymous forums messages are short, e.g. a 4chan message contains 15 words on average [3] and there are not more than 140 symbols for a Twitter post. A huge number of papers has been published on this and adjacent topics in recent years. Cerrea et al. studied the influence of complete anonymity on the users' behavior [6] in comparison with partial anonymity of Twitter. They have found that users tend to be more open and are more ready to express negative emotions (not only aggression) in anonymous environment. However, they have studied a site Whisper designed to share secrets and confessions, and it may influence their results. Martínez-Cámara has conducted an overview of different methods for Twitter sentiment analysis [7]. Another research was done by Dos Santos who successfully (from 76% to 88% for various measurement sets) detected the sentiment for Twitter messages [8] without using any handcrafted features, but unlike us he had labeled data. Tang and Wei analyzed Twitter sentiments using emoticons, smileys and neural networks [9].As we see, many modern researches use machine learning and neural networks methods for sentiment detection. However, Paltoglou [10] asserts that 'unsupervised' dictionary-based methods outperform 'state-of-the-art' machine learning. Nevertheless, he does not mention any deep learning or neural network-based methods, and his results are difficult to apply to other languages, besides English.

## 1. Methods and materials

Our study is focused on automatic identification of aggression for Russian and American imageboards. We have chosen 2ch.hk and 4chan.org as the most prominent and popular imageboards for their respective countries [11].

Aggression was detected by our algorithm based on the neural network algorithm word2vec [12] and its Gensim [13] implementation for the Python programming language. Word2vec is an unsupervised algorithm that allows finding semantic relations and distances between words without any annotation or other data preprocessing. Nowadays this algorithm is considered to be the best among other algorithms that allow finding semantic relations between words [14]. Although, some researchers argued that their methods performed better, for example, J. Pennington and R. Socher offered their method called GloVe (Global Vectors for Word Representations) and proved that their method outperforms word2vec [15]. However, other researchers found that word2vec is better in majority of cases and not so computationally expensive (a quadratic difference) [16].

First of all, we prepared the data for a more efficient training of a word2vec neural network. For training we used 654,047 4chan.org messages and 1,148,692 2ch.hk messages. We removed stopwords based on nltk-toolkit stopword list [17]. Then we turned tokens to their types for the Russian and English language using Snowball stemmer from nltk library. The stemmer doesn't consider context while prescribing type to a token but we needed to boost individual word occurrences to train our model more efficiently. Moreover, word2vec analyses the context of every word. After that we found set phrases for pairs of words that occur in some contexts significantly more often than in other. We did it with the help of Gensim package. With the help of word2vec neural net algorithm we trained two models, one for each analyzed imageboard. Then we built an automatic scikit-learn [18] implemented random forest classifier which we trained on manually annotated corpus of about 1000 messages (1308 messages for Russian, 1027 for English). 90% of messages were used for training and 10% for later evaluation. Two professionally trained linguists participated in the annotation. Basing on our word2vec model we used estimation of average semantic distance of words in the message to some manually chosen words and phrases that are typical for aggressive messages (we have chosen two obscene words for English and 5 for Russian) as parameters for our random forest classifier. We also estimated maximum and minimum semantic distance to our chosen words and between words in a message, as well as average, maximum and minimum length of words and messages. We tried using k-means clusterizing [19] to choose aggressive words automatically but we failed to get any advance with this method. After that we evaluated our classifiers' results and have found weights for most important features for the classification (Table. 1).

Table 1. Random forest feature importance

| Language (Imageboard) | Percentage of correct classification (%) | Weights of different parameters (Total – 1) | | | | | |
|---|---|---|---|---|---|---|---|
| | | Difference between maximum and minimum distance to the chosen wordset | Max closest distance to the chosen wordset | Average length of a word in message | Average distance of words in a message to the chosen wordset | Average semantic distance between words in a message | Other parameters |
| Russian (2ch.hk) | 59.13 | 0.158 | 0.156 | 0.037 | 0.158 | 0.150 | 0.341 |
| English (4chan.org) | 88.40 | 0.157 | 0.157 | 0.139 | 0.122 | 0.109 | 0.316 |

## 2. Discussion of results

The results for the English language are quite decent and after some additional testing and evaluation, the method may be used in practice.

Unfortunately, the results of automatic classification are very low for the Russian language. It may be connected with grammar and syntactic complexity of Russian. There are more tokens for one type form in Russian that is why the amount of annotated and not annotated data should be increased. We may also include some other information like parts of speech and other grammar characteristics as well as paradigmatic features, e.g. punctuation, emoticons, capitalization. Switching machine learning Random Forest algorithm to support-vector machines or some neural networks-based method may also help.

## 3. Conclusion

All in all we have gained a decent result for automatic aggression detection for the English language (88%). Yet our method may be still improved with the help of other features and parameters. However, the results of automatic aggression detection for the Russian language leave much to be desired (59%).

We are going to focus on considering and applying tagging of parts of speech and other grammatical characteristics of the text and in future research. We also would like to implement a doc2vec approach suggested by Le and Mikolov [20] or other similar method

**Acknowledgements**

The survey was partially funded by the Russian Science Foundation (RSF) in the framework of the project № 14-18-01059 at Institute of Applied and Mathematical Linguistics of the Moscow State Linguistic University (scientific head of the project – R. K. Potapova).